\documentclass[10pt,twocolumn,letterpaper]{article}

\usepackage[pagenumbers]{cvpr} 

\usepackage{graphicx}
\usepackage{amsmath}
\usepackage{amssymb}
\usepackage{booktabs, makecell, tabularx}
\usepackage[pagebackref,breaklinks,colorlinks]{hyperref}
\usepackage{algorithm}
\usepackage{algorithmicx}
\usepackage{algpseudocode}

\usepackage{enumitem}
\usepackage{adjustbox}
\usepackage{newfloat}
\usepackage[utf8]{inputenc}
\usepackage[T1]{fontenc}
\usepackage{listings}
\usepackage{bibentry}
\usepackage{arydshln}
\usepackage{pifont}
\usepackage[capitalize]{cleveref}
\usepackage[accsupp]{axessibility}

\crefname{section}{Sec.}{Secs.}
\Crefname{section}{Section}{Sections}
\Crefname{table}{Table}{Tables}
\crefname{table}{Tab.}{Tabs.}

\def\cvprPaperID{4078}
\def\confName{CVPR}
\def\confYear{2022}

\begin{document}

\newcommand\method{DiSparse}
\newcommand{\Hquad}{\hspace{0.5em}}
\newcommand{\cmark}{\ding{51}}%
\newcommand{\xmark}{\ding{55}}%
\definecolor{Gray}{gray}{0.9}
\definecolor{LightCyan}{rgb}{0.88,1,1}
\newcommand\ali[1]{\textcolor{blue}{Ali: #1}}

\def\ie{\emph{i.e}\onedot} \def\Ie{\emph{I.e}\onedot}

\title{DiSparse: Disentangled Sparsification for Multitask Model Compression}

\author{Xinglong Sun\textsuperscript{1}, Ali Hassani\textsuperscript{1}, Zhangyang Wang\textsuperscript{2}, Gao Huang\textsuperscript{3}, Humphrey Shi\textsuperscript{1,4}\\
{\small \textsuperscript{1}SHI Lab @ UIUC \& U of Oregon, \textsuperscript{2}UT Austin,\textsuperscript{3}Tsinghua University, \textsuperscript{4}Picsart AI Research (PAIR)}\\
}

\maketitle

\begin{abstract}
   Despite the popularity of Model Compression and Multitask Learning, how to effectively compress a multitask model has been less thoroughly analyzed due to the challenging entanglement of tasks in the parameter space. In this paper, we propose \textbf{\method{}}, a simple, effective, and first-of-its-kind multitask pruning and sparse training scheme. We consider each task independently by disentangling the importance measurement and take the unanimous decisions among all tasks when performing parameter pruning and selection. Our experimental results demonstrate superior performance on various configurations and settings compared to popular sparse training and pruning methods. Besides the effectiveness in compression, \method{} also provides a powerful tool to the multitask learning community. Surprisingly, we even observed better performance than some dedicated multitask learning methods in several cases despite the high model sparsity enforced by \method{}. We analyzed the pruning masks generated with \method{} and observed strikingly similar sparse network architecture identified by each task even before the training starts. We also observe the existence of a "watershed" layer where the task relatedness sharply drops, implying no benefits in continued parameters sharing. Our code and models will be available at: \href{https://github.com/SHI-Labs/DiSparse-Multitask-Model-Compression}{https://github.com/SHI-Labs/DiSparse-Multitask-Model-Compression}.
\end{abstract}

\section{Introduction}
\label{sec:intro}
Convolutional Neural Networks (CNNs)~\cite{lecun1998gradient} are considered the go-to architecture for computer vision, ever since the inception of AlexNet~\cite{krizhevsky2012imagenet}, especially in fundamental vision tasks such as image classification~\cite{deng2009imagenet}, object detection~\cite{liu2016ssd, he2017mask} and segmentation~\cite{long2015fully, chen2017deeplab}.
As more complex and difficult vision tasks are explored, substantial efforts are devoted to scaling deep convolutional networks to enormous sizes. Many models exist with parameters as many as billions, which significantly challenges those targeting edge device applications. Therefore, effectively compressing deep convolutional networks for efficient storage and computation has been a very active research area, and various approaches have been proposed and developed~\cite{cheng2017survey, deng2020survey} over the years.

Generally, neural network compression techniques can be categorized~\cite{deng2020survey} into pruning~\cite{lecun1990optimal, han2015deep, li2017pruning}, quantization~\cite{chen2015compressing, liu2018bi, yu2019any}, low-rank factorization~\cite{denton2014exploiting, zhang2015efficient, lin2018holistic}, and knowledge distillation~\cite{hinton2015distilling, jiao2019geometry, liu2019structured}. Network pruning, as a popular subfield of model compression, aims to discard certain parameters in the model, while retaining performance as much as possible. Pruning methods usually try to assign the best saliency score~(also referred to as importance score) to each parameter and perform selection and pruning based on these importance measurements. Despite the diversity in the pruning schemes proposed in recent years, chasing sparsity in the network, either in a structured or unstructured manner, has been one of the central themes since the very beginning~\cite{lecun1990optimal, hassibi1992second}. Parameter efficiency of sparse neural networks has been demonstrated~\cite{han2016eie, srinivas2017training}, and multiple works~\cite{park2016holistic, elsen2020fast} have shown inference time speedups are possible using sparsity for convolutional neural networks.

\begin{figure}[t]
  \centering
    \begin{subfigure}[b]{0.235\textwidth}
        \centering
        \includegraphics[width=\textwidth]{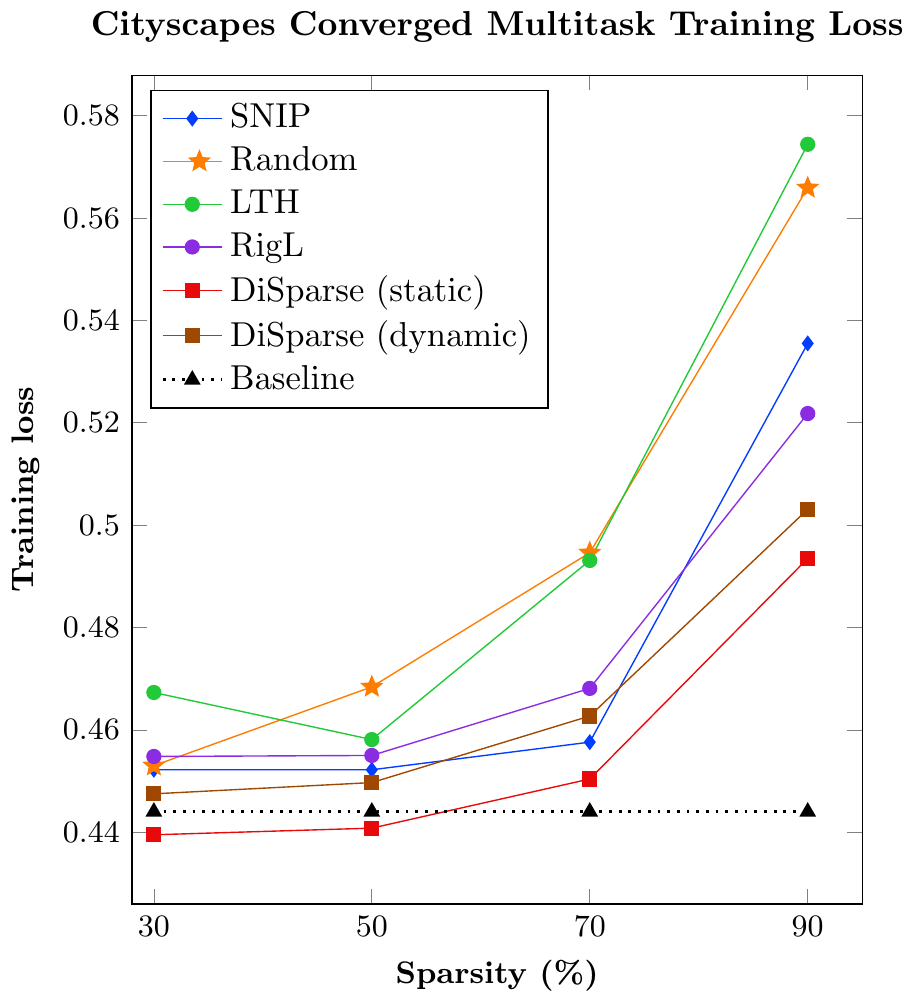}
    \end{subfigure}
    \hfill
    \begin{subfigure}[b]{0.235\textwidth}
        \centering
        \includegraphics[width=\textwidth]{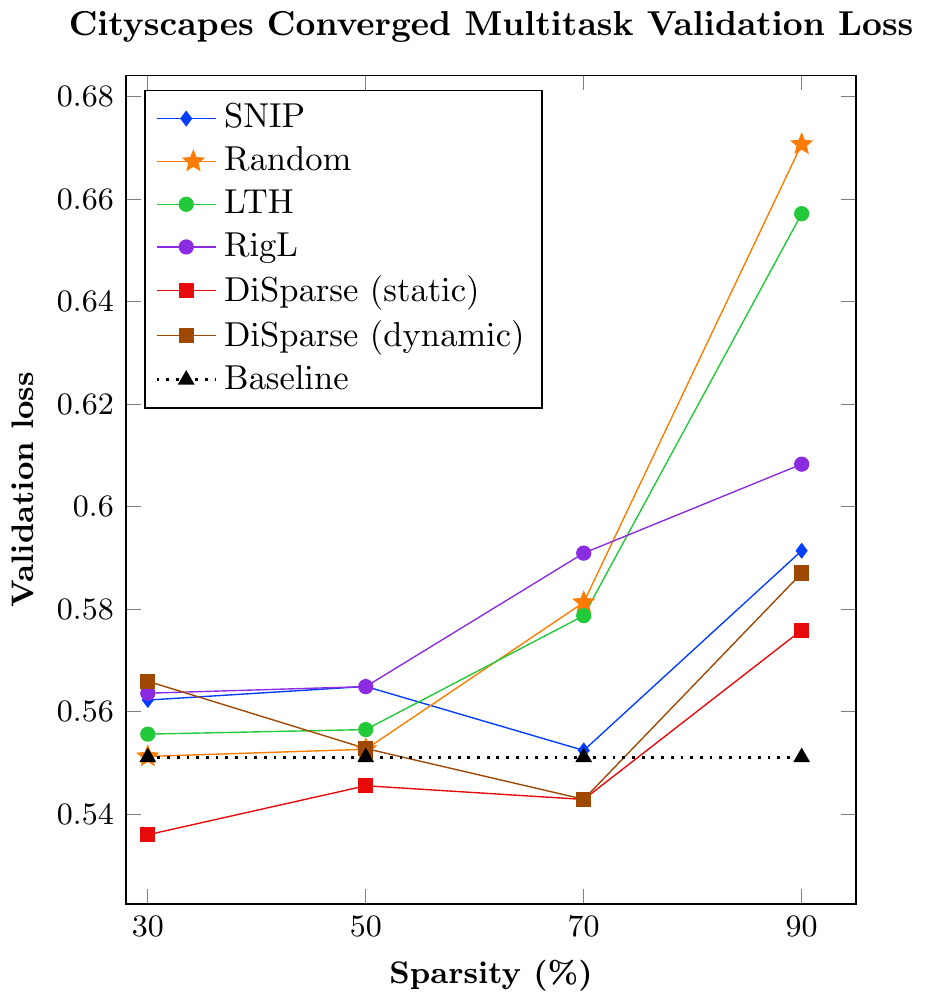}
    \end{subfigure}
   \caption{Converged training and validation multitask loss on Cityscapes. Our method DiSparse obtains the best training and validation behavior in both static and dynamic sparse training paradigms comparing to other methods. DiSparse even beats the unsparsified baseline at several sparsity levels.
}
   \label{fig:cityscape}
\end{figure}

Notwithstanding the increasing attention on model pruning and sparse training, effectively sparsifying a multitask network remains unexplored in spite of its importance. Multitask Learning (MTL) focuses on simultaneously solving multiple related tasks using a single model, which can significantly reduce the training and inference time and improve the generalization performance through learning a shared representation across tasks~\cite{caruana1997multitask}. It has a wide application in many problems like autonomous driving and indoor navigation robot where multiple tasks like semantic segmentation and depth estimation need to be performed simultaneously. A compact and efficient multitask network makes real-time performance possible on edge devices where computational resources are limited. Multitask network naturally brings storage and speed advantages over its single-task counterparts by a commonly shared backbone adopted in many popular MTL works~\cite{dvornik2017blitznet, jou2016deep, ranjan2017hyperface}. However, how to further compress and sparsify such networks hasn't been carefully analyzed. Compression on multitask networks with comparable performance as on single-task ones is very challenging because the shared space contains heavily entangled and intertwined features, causing the traditional pruning and sparse training algorithm to fail. A fraction of parameters in the shared space, though not important for one task, could be crucial for the performance of another. A few MTL works have explored the problem of entangled features and showed disentangling representation into shared and task-private spaces will improve the model performance ~\cite{liu2017adversarial, suteu2019regularizing}.

We propose the first-of-its-kind pruning scheme that enforces sparsity in multitask networks by taking the entangled nature of their features into consideration.
We argue that the key to properly compressing a multitask model is correctly identifying saliency scores for each task in the shared space, therefore \textbf{Spars}ifying in a \textbf{Di}sentangled manner (\textbf{\method{}}).
We take unanimous selection decisions among all tasks, which means that a parameter is removed only if it's shown to be not critical for any task.
This prevents extreme degradation in performance for certain tasks due to sparsification, leading to a more balanced network.

We conduct extensive experiments to validate the efficiency of our proposed scheme. We demonstrate compression performance on models of different structures with datasets of various sizes~\cite{silbermanECCV12, cordts2016cityscapes, zamir2018taskonomy}. We show results on pre-trained network, static sparse network at initialization, and dynamically growing sparse network at initialization. For each paradigm, we offered a slightly altered variant with the same core idea. We compared with popular pruning and sparse training methods and found that our proposed method is superior in both the training and validation phase, attaining lower training loss in a shorter period of time and better evaluation metrics on the validation set across different sparsity levels. Even more surprisingly, we observed better performance than some dedicated multitask learning approaches~\cite{misra2016cross, ruder2019latent, ahn2019deep} in several cases in spite of the high sparsity enforced in our model, showing the effectiveness of \method{} in multitask training. Besides the demonstration of superior model performance, we provide interesting observations in our experiments with \method{}. We compute the Intersection over Union (IoU) of the binary pruning masks generated with \method{} for each task to indicate task relatedness or similarity. Surprisingly, we observed strikingly similar sparse network architecture identified by each task even before the training starts. This offers a glimpse of the transferable subnetwork architecture across domains. Moreover, we observe the existence of a "watershed" layer where the task relatedness sharply drops, implying no benefits in continued parameters sharing. Exploitation of such property with \method{} could save tremendous labor and computation cost by obtaining a better multitask architecture pre-training. These observations show that \method{} does not only provide the compression community with the first-of-its-kind multitask sparsification scheme but also a powerful tool to the multitask learning community.

Our contributions can be summarized as follows:
\begin{itemize}
    \item Proposing a simple, effective, and first-of-its-kind pruning and sparse training scheme for multitask network by disentangling the importance measurements among tasks, leading to a more balanced network.
    \item Performing an extensive empirical study on multiple vision tasks and datasets, which demonstrates the superiority of \method{} compared to popular pruning and sparse training algorithms and even several dedicated multitask learning methods.
    \item Studying and discussing task relatedness and multitask model architecture design with \method{}, which provides a valuable tool to the multitask learning community from a compression perspective.
\end{itemize}
\section{Related Works}
\subsection{Pruning and Sparse Training}
Network pruning is effective in reducing inference cost and storage. Pruning can be roughly categorized into two categories by architecture: unstructured and structured pruning. Unstructured pruning methods~\cite{lecun1990optimal, han2015deep} drop less significant weights, regardless of where they occur. On the other hand, structured pruning methods~\cite{li2017pruning, liu2019structured}, operate under structural constraints, for example removing convolutional filters or attention heads~\cite{michel2019sixteen}, thus enjoy immediate performance improvement without specialized hardware or library support.
Pruning methods compute importance scores with different criterion to perform parameter selection. Most commonly used score criterion include: 1. Magnitude-based~\cite{han2015deep, li2017pruning}, 2. Gradient-based~\cite{molchanov2017variational, molchanov2019importance}, 3. Hessian-based~\cite{lecun1990optimal, hassibi1992second}, 4. Learning-based~\cite{liu2019structured, ding2021resrep}. Most pruning methods are only applied to pre-trained models.

Sparse training techniques, on the other hand, train sparse networks from scratch, and have also gained significant attention from the research community. The Lottery Ticket Hypothesis (LTH)~\cite{frankle2018lottery} hypothesized that if we can find a sparse neural network with iterative pruning, then we can train that sparse network from scratch to the same level of accuracy, by starting from the same initial conditions.
Single-Shot Network Pruning (SNIP)~\cite{lee2018snip} attempts to find an initial mask in a data-driven approach with one-shot pruning and uses this initial mask to guide parameters selection. As these methods still maintain a static network architecture throughout the training process, we refer to this group of methods as Static Sparse Training.

Another direction of sparse training is making network connections dynamic and allowing pruned weights to grow back. This allows for adaptive identification of high-quality sparse subnetworks. Sparse Evolutionary Training (SET)~\cite{mocanu2018scalable} prunes weights according to the standard magnitude criterion then adds weights back at random.
Evci~\etal~\cite{evci2020rigging} proposed Rigging the Lottery (RigL), an idea similar to SET but grows the weights back based on their gradients.
We refer to this group of methods as Dynamic Sparse Training due to the model's ability to dynamically grow back pruned weights.

\subsection{Multitask Network Compression}
Research in multitask network compression has recently gained attention. Several methods~\cite{he2021pruning, cheng2021multi, he2018multi} start with single-task networks and gradually merge them into a unified one, using feature sharing and similarity maximization. However, these schemes are inapplicable to pre-designed multitask models. This motivated our work, in which we propose a method that provides a pruning and sparse training scheme targeting a unified multitask network with shared parameters between tasks. To the best of our knowledge, our proposed method is the first to do so.

\section{Methodology}
\subsection{Notations}
\label{subsec:notations}
Given a dataset $\mathcal{D}$ with individual samples $x_i$, targets $y_i$, and a desired sparsity level $\mathcal{S} \in (0, 1)$ (\ie{} the ratio of zero weights), pruning or sparse training a neural network can be written as the following optimization problem:
\begin{align}
    \label{eqn:1}
    \min_\Theta L(\Theta; \mathcal{D}) &= \min_\Theta \frac{1}{n}\sum_{i=1}^{n} \ell(f(\Theta; x_i), y_i)\\
     \textrm{s.t.} \quad & \Theta \in \mathbb{R}^m, \quad \|\Theta\|_0 \leq (1-\mathcal{S})\cdot m \nonumber
\end{align}
Here, $\ell(.)$ is the standard loss function, $\Theta$ is the set of parameters of the neural network, $m$ is the total number of parameters in the model, and $\|.\|_0$ represents the $L_0$ norm.
For the ease of later notations, we reformulate the pruning and sparse training problem to find an optimal binary mask and modify Equation~(\ref{eqn:1}) as follows:
\begin{align}
    \label{eqn:2}
    \min_\mathcal{B} L(\Theta,\mathcal{B}; \mathcal{D}) &= \min_\mathcal{B} \frac{1}{n}\sum_{i=1}^{n} \ell(f(\Theta \odot \mathcal{B}; x_i), y_i)\\
     \textrm{s.t.} \quad  \Theta \in \mathbb{R}^m,& \mathcal{B} \in \mathbb{R}^m, \mathcal{B} \in \{0,1\}^m, \|\mathcal{B}\|_0 \leq (1-\mathcal{S})\cdot m \nonumber
\end{align}
where $\mathcal{B}$ represents a binary mask over the parameters, indicating which are kept and which are pruned.
Given a set of $\mathcal{K}$ tasks $\mathcal{T} = \{\mathcal{T}_1, \mathcal{T}_2, \hdots, \mathcal{T}_\mathcal{K}\}$, we represent the parameters used only by the $k^{th}$ task as $\Theta^{k} \in \mathbb{R}^{m_k}$ and the common parameters used by all of the tasks as $\Theta^{c} \in \mathbb{R}^{m_c}$. We also denote $\Theta^k \cup \Theta^c$ as $\Theta^{kc}$, and $\mathcal{B}^{kc}$ as its corresponding mask. Therefore masked parameters in the sparse network for task $\mathcal{T}_k$ can be expressed as $\Theta^{kc} \odot \mathcal{B}^{kc}$.
Furthermore, the target for $k^{th}$ task is denoted with $y^k_i$ for the $i^{th}$ data sample, as well as the loss function the same task with $\ell^k$. As a result, total loss for all tasks is expressed as follows:
\begin{align}
\label{eqn:3}
    \ell(f(\Theta; x_i), y_i) = \sum_k^\mathcal{K} \lambda^k \ell^k(f(\Theta^{kc} \odot \mathcal{B}^{kc}; x_i), y_i^k)
\end{align}
where $\lambda^k$ represents the weighting scalar for each task. Equation~(\ref{eqn:2}) can therefore be re-written as:

\begin{align}
    \min_\mathcal{B} L(\Theta,\mathcal{B}; \mathcal{D}) &= \min_\mathcal{B} \frac{1}{n}\sum_{i=1}^{n} \sum_k^\mathcal{K} \lambda^k \ell^k(f(\Theta^{kc} \odot \mathcal{B}^{kc}; x_i), y_i^k) \nonumber \\
    &= \min_\mathcal{B} \sum_k^\mathcal{K}L^k(\Theta^{kc} \odot \mathcal{B}^{kc}; \mathcal{D})\\
    \textrm{s.t.} \quad  \Theta \in \mathbb{R}^m,& \mathcal{B} \in \mathbb{R}^m, \mathcal{B} \in \{0,1\}^m, \|\mathcal{B}\|_0 \leq (1-\mathcal{S})\cdot m \nonumber
\end{align}

\subsection{Our Method}
\label{subsec:ours}
When solving for task-specific binary masks, $\mathcal{B}^k$, traditional pruning or sparse training methodologies work as proposed. However, while solving for $\mathcal{B}^c$, the binary mask for the large number of commonly shared parameters, we can't simply apply typical methods which directly utilize $L(\Theta,\mathcal{B}; \mathcal{D})$ as guidance, because the shared parameters are entangled with multiple tasks. Therefore, we propose a scheme for solving the shared mask, $\mathcal{B}^c$. For a given task $\mathcal{T}_k$, we compute the binary mask $\mathcal{B}^{kc}$ solely based on the task itself. This means saliency scores are computed solely based on $L^k(\Theta^{kc}, \mathcal{B}^{kc}; \mathcal{D})$. This mask will capture the preferred sub-network structure provided that we solve for task $\mathcal{T}_k$ independently. We denote the mask corresponding to the shared parameters in $\mathcal{B}^{kc}$ as $\mathcal{C}(\mathcal{B}^{kc}) \in \mathbb{R}^{m_c}$ and the task-private parameters in $\mathcal{B}^{kc}$ as $\mathcal{P}(\mathcal{B}^{kc}) \in \mathbb{R}^{m_k}$. We can then make the direct assignment $\mathcal{B}^k = \mathcal{P}(\mathcal{B}^{kc})$. To solve for the mask for the shared parameters, $B^c$, we feed all of the masks into an "arbiter" function, $\mathcal{A}$, and compute the final mask $\mathcal{B}^c$ for the shared parameters shown as follows:
\begin{align}
    \mathcal{B}^c = \mathcal{A}(\mathcal{C}(\mathcal{B}^{1c}), \mathcal{C}(\mathcal{B}^{2c}), \hdots, \mathcal{C}(\mathcal{B}^{\mathcal{K}c}))
\end{align}
We use three variants of saliency score criterion based on our proposed scheme for three different popular paradigms to enforce sparsity in the model: 1). Static Sparse Training 2). Dynamic Sparse Training 3). Pruning on pre-trained models.
In the following subsections, we will outline how $\mathcal{B}^{kc}$ is computed for each paradigm.

\begin{figure*}
\begin{center}
\includegraphics[width=0.80\linewidth]{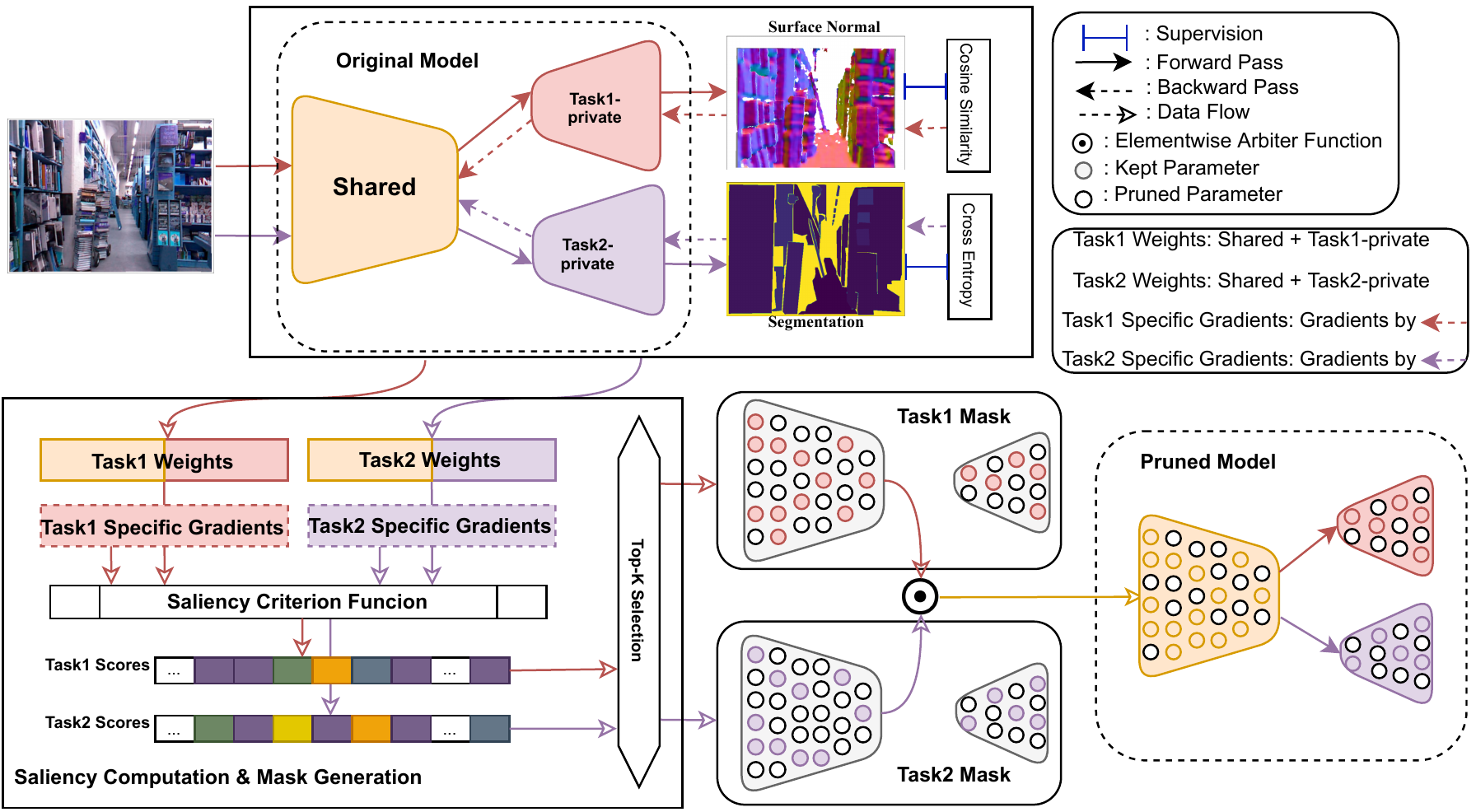}
\end{center}
  \caption{An overview of \method~. For task $\mathcal{T}_k$, we feed weights $\Theta^{kc}$ and their gradients w.r.t the loss $L^k(.)$ into a saliency scoring function to get their importance scores. Later we generate an optimal binary mask $\mathcal{B}^{kc}$ for the model assuming that we're only training the network independently for task $\mathcal{T}_k$. We directly assign $\mathcal{P}(\mathcal{B}^{kc})$, the task-private part, to $\mathcal{B}^{k}$ used as the pruning or growing mask for the task-private parameters. For $\mathcal{C}(\mathcal{B}^{kc})$, the shared part, we feed all of $\{\mathcal{C}(\mathcal{B}^{kc}), \forall k \in \{1,\hdots, \mathcal{T}\}\}$ to an element-wise arbiter function $\mathcal{A}$ and take its output as $\mathcal{B}^c$, the pruning or growing mask for the shared parameters.}
\label{fig:flowchart}
\end{figure*}

\begin{algorithm}[t]
    \caption{Proposed Algorithm: \method{}}\label{euclid}
    \label{algo:1}
    \textbf{Input:} $\Theta^{kc} \in \mathbb{R}^{m_k+m_c}, L^k(.) \quad \forall k \in \{1, 2, \hdots, \mathcal{K}\}$\\
    $\mathcal{A}, \mathcal{S}, \mathcal{D}, saliency(.)$\\
    \textbf{Output:} $\mathcal{B}$
    \begin{algorithmic}[1]
        \State{$\mathcal{B}cs \gets [\Hquad]$}
        \For{$k \gets 1$ to $\mathcal{K}$}
            \State{$\Vec{sk} \gets saliency(\Theta^{kc}, L^k(.), \mathcal{D})$}
            \State{$\gamma \gets (1-S)\cdot(m_k+m_c)$}
            \State{$\Tilde{sk\gamma} \gets {Top}\gamma(\Vec{sk})$}
            \State{$\mathcal{B}kc \gets \Vec{sk} > \Tilde{sk\gamma}$} \hfill\texttt{\footnotesize//Element-wise Comparison}
            \State{$\mathcal{B}k \gets \mathcal{P}(\mathcal{B}kc)$}
            \hfill\texttt{\footnotesize//Task-specific Mask}
            \State{$\mathcal{B}cs.append(\mathcal{C}(\mathcal{B}kc))$}
            \hfill\texttt{\footnotesize//Common Mask}
        \EndFor
        \State{$\mathcal{B}c \gets \mathcal{A}(\mathcal{B}cs)$}
        \State{$\mathcal{B} \gets \mathcal{B}c \cup (\bigcup_{k=1}^{\mathcal{K}}\mathcal{B}k)$}\\
        \Return $\mathcal{B}$
    \end{algorithmic}
\end{algorithm}

\subsubsection{Static Sparse Training}
\label{subsubsec: static}
We borrowed the effective data-driven sensitivity measurement from SNIP~\cite{lee2018snip} and made modifications based on our scheme. We represent the saliency for task $\mathcal{T}_k$ of the $j^{th}$ weight in the parameter space $\Theta^{kc}$, $\Theta^{kc}_j$, as $s_j^{kc}$, which is computed as follows:

\begin{align}
    g_j^{kc}(\Theta^{kc}; \mathcal{D}) &= \frac{\partial L^k(\Theta^{kc}\odot \mathcal{B}^{kc}; \mathcal{D})}{\partial \mathcal{B}^{kc}_j}\\
    s_j^{kc} &= \frac{|g_j^{kc}(\Theta^{kc}; \mathcal{D})|}{\sum_z^{m_c}|g_z^{kc}(\Theta^{kc}; \mathcal{D})|}
\end{align}
We can then use the saliency measurements to compute the mask $\mathcal{B}^{kc}$ for task $\mathcal{T}_k$.
\begin{align}
    \gamma &= (1- S)\cdot (m_k + m_c)\\
    \mathcal{B}^{kc}_j &= \mathbb{I}[s_j^{kc} - \Tilde{s_\gamma^{kc}} \geq 0], \quad \forall j \in \{1\hdots m_k+m_c\}
\end{align}
Here, $\Tilde{s_\gamma^{kc}}$ is the $\gamma^{th}$ largest element in the vector $\Vec{s^{kc}}$.\\ We can thus use $\mathcal{B}^{kc}$ as discussed in~\ref{subsec:ours} to compute the final mask for the entire model.

\subsubsection{Dynamic Sparse Training}
\label{subsubsec:dynamic}
We follow RigL~\cite{evci2020rigging} in terms of the pruning and growing criterion and made modifications based on our scheme. For pruning, we performed the standard magnitude pruning. For growing, as in~\ref{subsubsec: static}, we compute the saliency for task $\mathcal{T}_k$ of the $j^{th}$ weight in the parameter space, $s_j^{kc}$, as follows:
\begin{align}
    g_j^{kc}(\Theta^{kc}; \mathcal{D}) &= \frac{\partial L^k(\Theta^{kc}; \mathcal{D})}{\partial \Theta^{kc}_j}\\
    s_j^{kc} &= |g_j^{kc}(\Theta^{kc}; \mathcal{D})|
\end{align}
We can then use the saliency measurements to compute the growing mask $\mathcal{B}^{kc}$ for task $\mathcal{T}_k$.
\begin{align}
    \gamma &= f_{decay}(t; \alpha, T_{end})\cdot(1-S)\cdot (m_k+m_c)\\
    \mathcal{B}^{kc}_j &= \mathbb{I}[s_j^{kc} - \Tilde{s_\gamma^{kc}} \geq 0], \quad \forall j \Hquad{} \textrm{s.t.} \Hquad{} \Theta^{kc}_j \not\in \Theta^{kc}\setminus \mathbb{I}_{active}
\end{align}
Here, $f_{decay}(.)$ is a decaying function as in RigL~\cite{evci2020rigging} to update the ratio of parameters to prune and grow at each update iteration, and $\Theta^{kc}\setminus \mathbb{I}_{active}$ is the set of active connections remaining after pruning.\\ We can thus use $\mathcal{B}^{kc}$ as discussed in~\ref{subsec:ours} to compute the final mask for the entire model.

\subsubsection{Pruning on Pre-trained Models}
\label{subsubsec:pretrained}
We borrowed the pruning criterion from GF~\cite{lubana2020gradient} and made modifications based on our scheme. As in~\ref{subsubsec: static}, we compute the saliency for task $\mathcal{T}_k$ of the $j^{th}$ weight in the parameter space, $s_j^{kc}$, as follows:
\begin{align}
    g_j^{kc}(\Theta^{kc}; \mathcal{D}) &= \frac{\partial L^k(\Theta^{kc}; \mathcal{D})}{\partial \Theta^{kc}_j}\\
    s_j^{kc} &= |g_j^{kc}(\Theta^{kc}; \mathcal{D})|\cdot |\Theta^{kc}_j|^2
\end{align}
Therefore, we can use the saliency measurements to compute the pruning mask $\mathcal{B}^{kc}$ for task $\mathcal{T}_k$.
\begin{align}
    \gamma &= (1- S)\cdot (m_k + m_c)\\
    \mathcal{B}^{kc}_j &= \mathbb{I}[s_j^{kc} - \Tilde{s_\gamma^{kc}} \geq 0], \quad \forall j \in \{1\hdots m_k+m_c\}
\end{align}
\\ We can thus use $\mathcal{B}^{kc}$ as discussed in~\ref{subsec:ours} to compute the final mask for the entire model.
\subsubsection{Arbiter Function}
\label{subsubsec:arbiter}
As discussed in~\ref{subsec:ours}, we used an "arbiter" function, $\mathcal{A}$, to solve the mask for the common parameters, $\mathcal{B}^c$. We provided two choices here:

\noindent\textbf{Element-wise Logical OR. }
We performed a logical OR operation over the mask computed by each task. In this case, a parameter will be pruned only if it's considered not important for each task.
\begin{align}
    \mathcal{B}^c &= \mathcal{A}(\mathcal{C}(\mathcal{B}^{1c}), \mathcal{C}(\mathcal{B}^{2c}) \hdots\mathcal{C}(\mathcal{B}^{\mathcal{K}c})) \nonumber\\
    &= \mathcal{C}(\mathcal{B}^{1c}) | \mathcal{C}(\mathcal{B}^{2c}) | \hdots | \mathcal{C}(\mathcal{B}^{\mathcal{K}c})
\end{align}

\noindent\textbf{Majority Vote. }
For three or more tasks~($\mathcal{K} \geq 3$), element-wise majority vote can be applied for more effective compression. A parameter will be pruned if most of the tasks agree to remove this particular connection.
\begin{align}
    \mathcal{B}^c &= \mathcal{A}(\mathcal{C}(\mathcal{B}^{1c}), \mathcal{C}(\mathcal{B}^{2c}) \hdots\mathcal{C}(\mathcal{B}^{\mathcal{K}c})) \nonumber\\
    &= MAJ(\mathcal{C}(\mathcal{B}^{1c}), \mathcal{C}(\mathcal{B}^{2c}) \hdots\mathcal{C}(\mathcal{B}^{\mathcal{K}c}))
\end{align}

\section{Empirical Evaluation}

\begin{table*}[ht!]
    \centering
    \resizebox{0.925\textwidth}{!}
    {
        \begin{tabular}{l|cc|ccccc|c|c}
            \toprule
            Model  & \multicolumn{2}{c}{T1: Semantic Seg.} & \multicolumn{5}{c}{T2: SN Prediction} & Sparsity & Pre-trained\\
            & mIoU$\uparrow$ & PixelAcc$\uparrow$ & Mean Err$\downarrow$ & Median Err$\downarrow$ & 11.25$\uparrow$ & 22.5$\uparrow$  & 30$\uparrow$& (\%)$\uparrow$\\
            \midrule
            DeepLab~\cite{chen2017deeplab}{\footnotesize (baseline)} & 27.69 & 58.77 & 16.55 & 14.17 & 39.62 & 73.54 & 86.33 & 0 & N/A \\
            LTH~\cite{frankle2018lottery} & 24.63 & 56.25 & 17.01 & 14.27 & 39.49 & 71.95 & 84.31 & 90.00 & \cmark\\
            SNIP~\cite{lee2018snip} & 23.83 & 56.90 & 16.58 & 14.05 & 39.82 & 73.73 & 85.77 & 90.00   & \xmark\\
            Random & 25.56 & 25.18 & 18.86 & 16.22 & 27.81 & 69.55 & 85.45 & 90.00 & \xmark\\
            \method{} (\textit{Ours}) & \textbf{26.48} & \textbf{57.77} & \textbf{16.44} & \textbf{13.69} & \textbf{41.24} & \textbf{74.07} & \textbf{85.85} & 90.00 &\xmark\\
            \midrule
            RigL~\cite{evci2020rigging} & 24.91 & 56.74 & 17.27 & 14.63 & 38.52 & 70.75 & 83.71 & 90.00 &\xmark\\
            \method{} (\textit{Ours}) & \textbf{28.16} & \textbf{59.18} & \textbf{16.54} & \textbf{13.47} & \textbf{42.25} & \textbf{73.14} & \textbf{84.73} & 90.00 &\xmark\\
            \midrule
            IMP~\cite{han2015deep} & 29.19 & 59.81 & 16.58 & 13.32 & 43.32 & 72.31 & 84.03 & 90.00 &\cmark\\
            Random & 25.56 & 25.18 & 18.69 & 16.04 & 23.37 & \textbf{73.08} & \textbf{86.78} & 90.00 &\cmark\\
            \method{} (\textit{Ours}) & \textbf{29.45} & \textbf{59.95} & \textbf{16.56} & \textbf{13.30} & \textbf{43.33} & 72.49 & 84.18 & 90.00 &\cmark\\
            \bottomrule
        \end{tabular}
    }
    \caption{\method{} semantic segmentation and surface normal prediction results on NYU-v2~\cite{silbermanECCV12} compared to static sparse training, dynamic sparse training, and pre-trained model pruning methods.}
    \label{table:nyu}
\end{table*}

We conduct extensive experiments to show that our proposed method outperforms many strong baselines and is very effective in sparsifying a multitask model.
\subsection{Datasets, Tasks, and Model}
We evaluate our method on three popular multitask datasets: NYU-v2~\cite{silbermanECCV12}, Cityscapes ~\cite{cordts2016cityscapes}, and Tiny-Taskonomy~\cite{zamir2018taskonomy}. We perform joint Semantic Segmentation and Surface Normal Prediction~\cite{misra2016cross, gao2019nddr, sun2019adashare} for NYU-v2, and joint Semantic Segmentation and Depth Prediction~\cite{sun2019adashare} for Cityscapes. For Tiny-Taskonomy, we perform joint training on 5 tasks: Semantic Segmentation, Surface Normal Prediction, Depth Prediction, Keypoint Detection, and Edge Detection~\cite{sun2019adashare, standley2020tasks}.

We performed sparsification on the widely-used DeepLab-ResNet~\cite{chen2017deeplab} with atrous convolutions as the backbone and the ASPP~\cite{chen2017deeplab} architecture for task-specific heads. As discussed in the above sections, we use the popular multitask architecture, in which all tasks share the backbone but have separate task-specific heads.

\subsection{Evaluation Metrics}
\label{subsec:metrics}
We demonstrated sparsity level as the most direct evaluation metric for the pruned or the sparse-trained model. The higher the sparsity level of a model, the more zeros it contains thus enjoying more acceleration and memory benefits. In terms of the performance evaluation of tasks, for Semantic Segmentation, we show mean Intersection over Union (mIoU) and Pixel Accuracy (Pixel Acc). For Surface Normal Prediction, we use mean and median angle error as the main evaluation metrics. On NYU-v2, we also demonstrate the percentage of pixels whose prediction is within the angles of $11.25^\circ$, $22.5^\circ$, and $30^\circ$ to the ground truth~\cite{eigen2015predicting}. For Depth Prediction, we compute absolute and relative errors as the main evaluation metrics. On Cityscapes, we also show the relative difference between the prediction and ground truth via the percentage of $\delta = max(\frac{y_{pred}}{y_{gt}}, \frac{y_{gt}}{y_{pred}})$ within threshold $1.25$, $1.25^2$, and $1.25^3$~\cite{eigen2014depth}. In both keypoints and edge detection tasks, we choose to show the mean absolute error to the provided ground-truth map as the main evaluation metric. Since for multitask learning, it's hard to qualify the performance of a model or algorithm with a single metric, we also demonstrate the converged training and testing multitask loss to show how our proposed scheme helps the model to learn in the training phase.

\subsection{Experimental Settings}
\label{subsec:exp_set}
We used PyTorch for all of our experiments, and two RTX 2080 Ti GPUs.
We used Adam optimizer and a batch size of $16$ for all experiments. We trained NYU-v2 for $20K$ iterations and used an initial learning rate of $1e-3$, decaying by $0.5$ every $4000$ iterations. We used the same optimization settings for Cityscapes, except for the initial learning rate, which was set to $1e-4$. Tiny-Taskonomy was trained for $100$K iterations with an initial learning rate of $1e-4$ decaying by $0.3$ every $12$K iterations.
We used cross-entropy loss for Semantic Segmentation, negative cosine similarity between the normalized prediction and ground-truth for Surface Normal Prediction, and L1 loss for the rest of the tasks.
To avoid bias and diversity in different pre-trained models, we train all of the models from scratch for a fair comparison among different methods.

\begin{table*}[]
    \centering
    \resizebox{0.925\textwidth}{!}{
        \begin{tabular}{l|ccc|ccccc|c|c}
            \toprule
            Model  & \multicolumn{3}{c}{T1: Semantic Seg.} & \multicolumn{5}{c}{T2: Depth Prediction} & Sparsity & Pre-trained\\
            & mIoU~$\uparrow$             & PixelAcc~$\uparrow$ & Error~$\downarrow$       & Abs. Error~$\downarrow$ & Rel. Error~$\downarrow$ & $\delta 1.25$ $\uparrow$ & $\delta 1.25^2$ $\uparrow$  & $\delta 1.25^3$ $\uparrow$    & (\%) $\uparrow$      \\
            \midrule
            DeepLab~\cite{chen2017deeplab}{\footnotesize (baseline)} & 42.58 & 74.84 & 0.49 & 0.016 & 0.33 & 74.22 & 88.90 & 94.47 & 0 & N/A\\
            LTH~\cite{frankle2018lottery} & 36.19 & 73.97 & 0.51 & 0.017   & \textbf{0.35} & 72.46 & 87.29 & 93.59 & 90.00   & \cmark \\
            SNIP~\cite{lee2018snip} & \textbf{38.47} & 73.99& 0.53   & 0.016   & 0.36        & 72.49 & 87.51 & \textbf{93.61} & 90.00   & \xmark\\
            Random & 35.67& 71.74& 0.60   & 0.021   & 0.43 & 66.47 & 82.21 & 90.63 & 90.00   & \xmark \\
            \method{} (\textit{Ours}) & 38.21 & \textbf{74.31} & \textbf{0.49}  & \textbf{0.015}   & \textbf{0.35} & \textbf{72.72} & \textbf{87.60} & 93.59 & 90.00   & \xmark\\
            \midrule
            RigL~\cite{evci2020rigging} & 36.57 & 74.02 & 0.51 & 0.017   & 0.37 & 71.98 & 86.66 & 93.05 & 90.00   & \xmark   \\
            \method{} (\textit{Ours}) & \textbf{38.55} & \textbf{74.28} & \textbf{0.49}   & \textbf{0.016}   & \textbf{0.36}        & \textbf{72.11} & \textbf{87.01} & \textbf{93.41} & 90.00   & \xmark   \\
            \midrule
            IMP~\cite{han2015deep} & 37.63  & 73.90 & 0.50   & 0.018   & \textbf{0.37} & 71.15 & 86.41 & 93.01 & 90.00   & \cmark \\
            Random & 33.62 & 51.63 & 1.18   & 0.034   & 0.43 & 50.09 & 72.42 & 82.73 & 90.00   & \cmark \\
            \method{} (\textit{Ours}) & \textbf{40.71} & \textbf{74.64} & \textbf{0.49} & \textbf{0.016}   & \textbf{0.37}        & \textbf{72.06} & \textbf{86.67} & \textbf{93.11} & 90.00   & \cmark \\
            \bottomrule
        \end{tabular}
    }
    \caption{\method{} semantic segmentation and depth prediction results on Cityscapes~\cite{cordts2016cityscapes} compared to static sparse training, dynamic sparse training, and pre-trained model pruning methods.}
    \label{table:city}
\end{table*}

\subsection{Baselines and Ours}
We evaluated several methods: LTH~\cite{frankle2018lottery}, RigL~\cite{evci2020rigging}, SNIP~\cite{lee2018snip}, and IMP~\cite{han2015deep}. We used SNIP's official implementation in Tensorflow, and implemented the rest in PyTorch.
For LTH~\cite{frankle2018lottery}, we used the fully-trained model to get the sub-network structure and rewind the model to the initial weights to start the sparse training. For SNIP~\cite{lee2018snip}, we use the gradients on $50$ random data batches drawn from the training datasets to compute the sparse mask. For RigL~\cite{evci2020rigging}, we followed the original paper and utilized a cosine decay to gradually decrease the pruning and growing ratio at each iteration. We stopped the update at the $75\%$-$th$ iteration. Also, for the initial sparsity distribution over layers,  we adopted \textit{Erd\H{o}s-R\'enyi-Kernel (ERK)} introduced in~\cite{mocanu2018scalable, evci2020rigging} as it was shown to attain the best performance.\\
For our method, we used the same configurations as SNIP in static sparse training experiments and RigL in dynamic sparse training ones.
In terms of the optimization, for the sparse training methods, we used the same settings as the original models described in~\ref{subsec:exp_set}. For the pruning on pre-trained model experiments, we retrained the model for $1000$ iterations with learning rate set as $1e-5$ after pruning for NYU-v2 and Cityscapes. For Tiny-Taskonomy, we retrained for $4000$ iterations with learning rate set as $1e-6$ decaying by $0.3$ every $1500$ iterations. Also, we select the Element-wise OR as our arbiter function for results reported in the tables but also explore the Majority Vote choice in the ablation studies.

\subsection{Quantitative Results}
We demonstrated results in three different learning scenarios and datasets in Table~\ref{table:nyu},~\ref{table:city}, and~\ref{table:taskonomy}. We performed all of the experiments on four different sparsity levels ($30\%, 50\%, 70\%, 90\%$) for all of the methods. Due to the space constraint, we only report evaluation results at the most extreme sparsity level, $90\%$. However, we do show the converged training and validation loss for all of the sparse training methods including two of ours at all sparsity levels on Cityscapes in Figure~\ref{fig:cityscape}. From the figure, the superiority of our proposed scheme is clearly observed in both the training and validation set. It attains much lower training loss across all sparsity levels than other methods and generalizes well on the validation set. For full evaluation results at all sparsity levels, please refer to the Appendix. In the tables, we reported all the evaluation metrics discussed in~\ref{subsec:metrics}. We also indicate whether the method utilized any information from the pre-trained model. Though LTH~\cite{frankle2018lottery} proposes a sparse training approach, it relies on the pre-trained model to extract the sub-network architecture. As shown in~\cite{zhou2019deconstructing}, the "winning lottery tickets" obtain non-random accuracies even before training has started. Thus, we also check the "pre-trained" mark for LTH in our tables. For all the learning scenarios, we also include random pruning or sparse training results.

On NYU-v2, as shown in Table~\ref{table:nyu}, \method{} outperforms all the other methods by a significant margin in all paradigms. In the static sparse training paradigm, \method{} gets really close overall performance to the original unsparsified model and even better performance on three metrics (\ie{} Median Angle Error, $11.25^\circ$, and $22.5^\circ$). As discussed in~\ref{subsec:ours}, we borrowed the saliency measurement from SNIP~\cite{lee2018snip} for static sparse training. As seen in the table, \method{} is much more effective compared to SNIP. In the dynamic training paradigm, \method{} performs much better than RigL~\cite{evci2020rigging}. Surprisingly, our performance is even better than the original unsparsified model by a noticeable margin, demonstrating the efficacy of our proposed scheme. In fact, the performance of \method{} surpasses several dedicated multitask learning methods~\cite{misra2016cross, ruder2019latent, ahn2019deep} in the same optimization settings but with zero sparsity. Comparison with such methods is included in the Appendix. On the paradigm where pruning is performed on pre-trained models, \method{} is also better than both IMP~\cite{han2015deep} and random pruning by a noticeable margin.

On Cityscapes, as shown in Table~\ref{table:city}, \method{} is shown to be more superior than other methods consistently across all paradigms. On static sparse training, though SNIP obtains better mIoU and close Pixel Acc to \method{}, it performs poorly on the depth prediction task. As another indicator of Semantic Segmentation task, we also include the mean absolute error indicated as "Error" in Table~\ref{table:city}. As shown, \method{} even obtains the same error as the baseline. On the dynamic sparse training and pruning paradigms, \method{} outperforms other methods by a substantial performance margin across all metrics.

\method{} is also superior in Tiny-Taskonomy, as shown in Table~\ref{table:taskonomy}. Its performance is demonstrated to be more balanced across tasks. For example, although SNIP gets slightly better mIoU and Pixel Acc, it performs even much worse than the random sparse training on surface normal prediction. Whereas our method is demonstrated to have decent performance on all of the 5 tasks and does not have one task being drastically worse than others. On the dynamic sparse training and pruning paradigms, \method{} is also shown to be much more effective than other methods.

The consistent superiority of \method{} on all three datasets shows that it can indeed help multitask model achieve high sparsity with minimum performance drop.

\begin{table*}[]
\centering
\resizebox{.925\textwidth}{!}{
\begin{tabular}{l|cc|cc|cc|c|c|c|c}
\toprule
Model  & \multicolumn{2}{c}{T1: Semantic Seg.} & \multicolumn{2}{c}{T2: SN Prediction} & \multicolumn{2}{c}{T3: Depth} & T4: Keypt. & T5: Edge & Sparsity & Pre-trained\\
& mIoU~$\uparrow$ & PixelAcc~$\uparrow$ & Mean Err.~$\downarrow$ & Med. Err.~$\downarrow$ & Abs. Err~$\downarrow$ & Rel. Err~$\downarrow$  & Err~$\downarrow$    &  Err~$\downarrow$ & (\%)~$\uparrow$ & \\
\midrule
DeepLab~\cite{chen2017deeplab}{\footnotesize (baseline)} & 29.51 & 92.98 & 24.36 & 12.88 & 0.021 & 0.033 & 0.20 & 0.21 & 90.00 & N/A \\
LTH~\cite{frankle2018lottery} & 24.36            & 91.62              & 26.07      & 15.78        & 0.023 & 0.036 & \textbf{0.20} & 0.22 & 90.00 & \cmark          \\
SNIP~\cite{lee2018snip} & \textbf{26.03}            & \textbf{91.89}              & 27.06      & 18.12        & \textbf{0.022} & 0.036 & \textbf{0.20} & \textbf{0.21} & 90.00 & \xmark          \\
Random & 25.39            & 91.18              & 26.31      & 16.30        & 0.023 & 0.037 & \textbf{0.20} & 0.22 & 90.00 & \xmark          \\
\method{} (\textit{Ours}) & 25.53            & 91.85              & \textbf{25.42}      & \textbf{14.80}        & \textbf{0.022} & \textbf{0.035} & \textbf{0.20} & \textbf{0.21} & 90.00 & \xmark          \\
\midrule
RigL~\cite{evci2020rigging} & 23.87            & 90.14              & 26.91      & 17.75        & 0.023 & 0.037 & \textbf{0.19} & \textbf{0.20} & 90.00 & \xmark \\
\method{} (\textit{Ours}) & \textbf{25.21}            & \textbf{91.49}              & \textbf{25.28}      & \textbf{14.63}        & \textbf{0.022} & \textbf{0.034} & \textbf{0.19} & \textbf{0.20} & 90.00 & \xmark \\
\midrule
IMP~\cite{han2015deep} & 26.71            & 91.74              & 24.99      & 13.89        & 0.021 & 0.033 & 0.20 & 0.21 & 90.00 & \cmark \\
Random & 23.89           & 83.89              & 41.21      & 38.35        & 0.25 & 0.39 & 0.45 & 0.40 & 90.00 & \cmark \\
\method{} (\textit{Ours}) & \textbf{29.12}            & \textbf{92.58}              & \textbf{24.70}      & \textbf{13.42}        & \textbf{0.021} & \textbf{0.033} & \textbf{0.19} & \textbf{0.20} & 90.00 & \cmark \\
\bottomrule
\end{tabular}}
\caption{\method{} semeantic segmentation, surface normal prediction, depth prediction, keypoint detection, and edge detection results on Tiny-Taskonomy~\cite{zamir2018taskonomy} compared to static sparse training, dynamic sparse training, and pre-trained model pruning methods.}
\label{table:taskonomy}
\end{table*}

\begin{table*}[]
    \centering
    \resizebox{0.95\textwidth}{!}
    {
        \begin{tabular}{l|cc|cc|cc|c|c|c|c}
        \toprule
        Model  & \multicolumn{2}{c}{T1: Semantic Seg.} & \multicolumn{2}{c}{T2: SN Prediction} & \multicolumn{2}{c}{T3: Depth} & T4: Keypt. & T5: Edge & Sparsity & Pre-trained\\
        & mIoU~$\uparrow$ & PixelAcc~$\uparrow$ & Mean Err.~$\downarrow$ & Med. Err.~$\downarrow$ & Abs. Err.~$\downarrow$ & Rel. Err.~$\downarrow$  & Err.~$\downarrow$ & Err.~$\downarrow$ & (\%)~$\uparrow$ \\
        \midrule
        \method{}~(static) & 26.03~(\textbf{0.50})            & 92.03~(\textbf{0.18})              & 25.28~(\textbf{-0.14})      & 14.52~(\textbf{-0.32})   & 0.022~(0.00) & 0.035~(0.00) & 0.20~(0.00) & 0.22~(0.01) & 90.00 & \xmark          \\
        \method{}~(dynamic) & 24.97~(-0.24)            & 91.77~(\textbf{0.28})              & 25.89~(0.61)      & 15.88~(1.25)       & 0.023~(0.001) & 0.036~(0.002) & 0.20~(0.01) & 0.21~(0.01) & 90.00 & \xmark \\
        \bottomrule
        \end{tabular}
    }
    \caption{Ablations on Tiny-Taskonomy~\cite{zamir2018taskonomy}. We show the results with the majority vote. Values inside the parenthesis are changes from the corresponding metrics in Table~\ref{table:taskonomy}. Positive changes are bold.}
    \label{table:ablation}
\end{table*}

\subsection{Ablation Studies}
We explored the effect of the arbiter function $\mathcal{A}$ and demonstrated the results in Table~\ref{table:ablation}. As discussed above, results reported in Table~\ref{table:nyu},~\ref{table:city}, and~\ref{table:taskonomy} come from methods with Element-wise OR as the arbiter function. Here, we tried using the Element-wise Majority Vote on Tiny-Taskonomy since it only makes sense when we have a relatively large number of tasks~($\mathcal{K} \geq 3$). Compared to results shown in Table~\ref{table:taskonomy}, Majority Vote yields better results in the static sparse training paradigm. It gives better performance on Semantic Segmentation and Surface Normal Prediction. However, in the dynamic sparse training setup, it doesn't surpass the performance of the Element-wise OR option but is still better than RigL in overall performance.

\subsection{Limitation}
As shown in~\ref{subsec:ours}, the sparsity of our model will not be exactly as requested due to the operations taken by the arbiter function $\mathcal{A}$.
For the dynamic sparse training experiments, since \method{} grows more weights in the growing session, we simply adjust the pruning rate in the following pruning session to compensate for it. For example, suppose the expected sparsity level at iteration $i$ is $S_i$ and the expected pruning rate is $P_i$. Our method generates a model with sparsity $\hat{S_i}, \hat{S_i} \leq S_i$. We adjust the pruning rate to $\hat{P_i} = 1 - \frac{1-S_i}{1-\hat{S_i}}\cdot(1-P_i)$ to keep the final sparsity as expected. For static sparse training, it’s more complex and we set up an empirical approach. We start from the requested sparsity and iteratively increase it and run \method{} until the final sparsity is within an acceptable margin. When the final sparsity is a little bit off, we always keep it higher than that of the baselines for a fair comparison.

\subsection{Discussion}

\begin{figure}[t]
  \centering
   \includegraphics[width=0.9\linewidth]{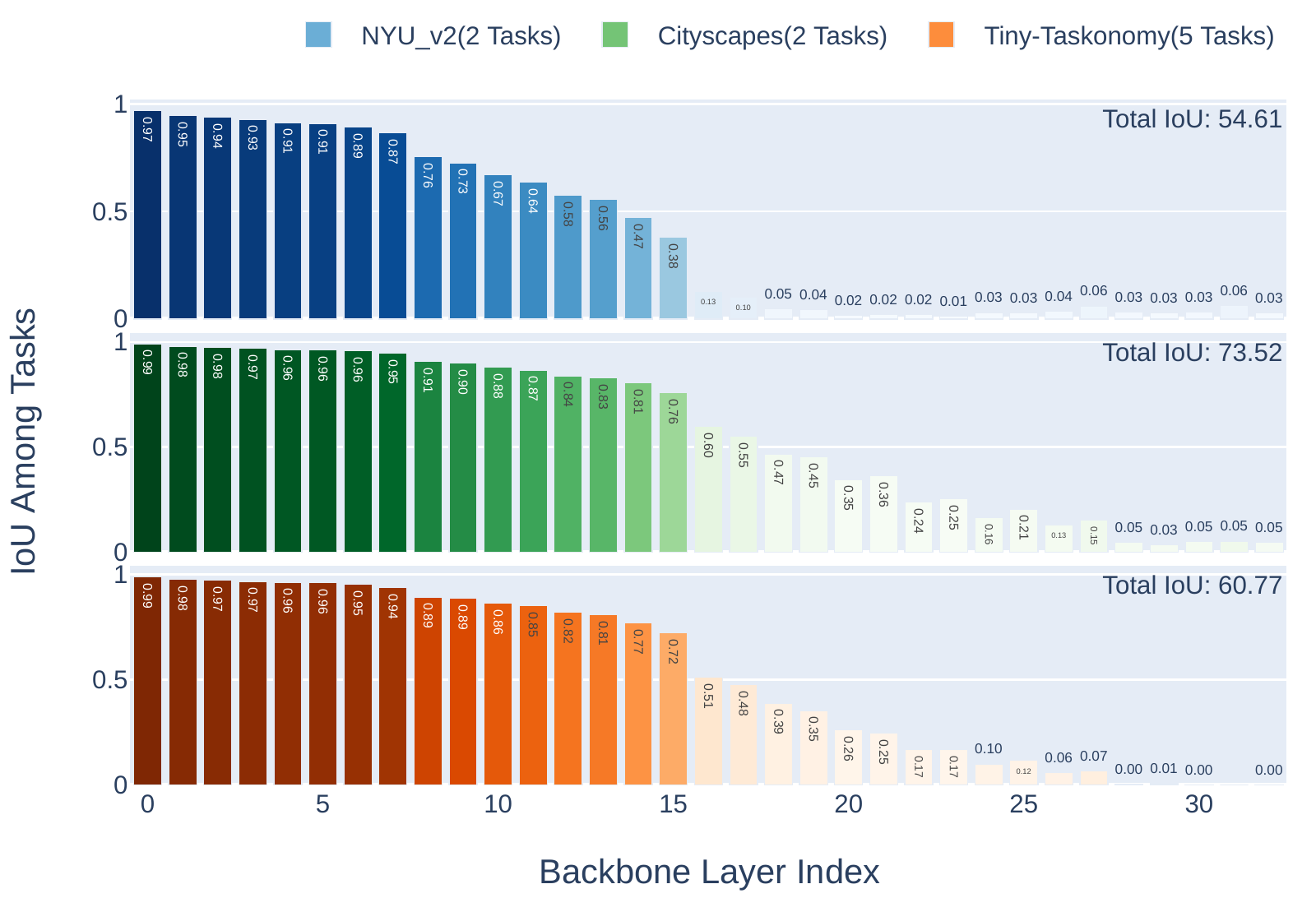}
   \caption{Layer-wise IoU evaluating preferred sparse subnetwork architecture among different tasks on three different datasets.}
   \label{fig:density}
\end{figure}

LTH~\cite{frankle2018lottery} inspired many to study the transferability of the "winning ticket", or sparse subnetwork architecture in general, across domains~\cite{morcos2019one, paganini2020bespoke, sabatelli2021transferability}.
Morcos~\etal~\cite{morcos2019one} showed that winning ticket initializations generalized across a variety of natural image datasets, suggesting that different tasks seem to enjoy the same sparse subnetwork structure.
Here, we made another exploratory step towards finding a transferable across domains sparse sub-network architecture. In the static sparse training setups, before the training even started, we analyzed the masks generated by different tasks regarding the commonly shared backbone with criterion discussed in ~\ref{subsubsec: static}. To evaluate the task similarity and relatedness, we compute Intersection over Union~(IoU) between their computed masks, which indicate the agreement and divergence of the preferred sparse subnetwork architecture by different tasks. Similar to ~\ref{subsec:ours}, we use $\mathcal{C}(\mathcal{B}^{kc})$ to denote the preferred mask on the shared backbone for task $\mathcal{T}_k$. IoU can therefore be computed as $IoU = \frac{|\bigcap_{k=1}^{\mathcal{K}}\mathcal{C}(\mathcal{B}^{kc})|}{|\bigcup_{k=1}^{\mathcal{K}}\mathcal{C}(\mathcal{B}^{kc})|}$. We present layer-wise IoU scores from the first 32 layers of the backbone on all three datasets in Figure~\ref{fig:density}. Surprisingly, we observed strikingly high IoU among tasks, even in the 5-task Tiny-Taskonomy dataset. This implies that even before training starts, different tasks tend to select the same architecture in the shared parameter space to facilitate training, suggesting potential for domain-independent sparse architecture exploration. Another interesting observation is that IoU drops sharply at certain layers. This phenomenon is observed across all three datasets, though this "watershed" layer is different (Layer 18 on NYU-v2, Layer 28 on Cityscapes, and Layer 26 on Taskonomy). This IoU analysis among tasks could be very helpful for pre-training multitask network design. For example, we could stop sharing parameters in the backbone and start branching out for different tasks at the layer where IoU sharply drops. We leave this for future discussions.

\section{Conclusion}
In this paper, we present \method{}, a novel sparse training and pruning method targeting multitask models. We conduct extensive experiments which demonstrate its superiority compared to other related methods on three different datasets and learning scenarios. We also provide promising future research directions in transferrable lottery-ticket across domains and multitask architecture design with our proposed scheme.


\clearpage
{\small
\bibliographystyle{ieee_fullname}
\bibliography{references}
}

\clearpage

\setlength{\tabcolsep}{4pt}
\begin{table*}[b!]
    \centering
    \resizebox{0.993\textwidth}{!}
    {
        \begin{tabular}{l|cc|ccccc|c|c}
            \toprule
            Model  & \multicolumn{2}{c}{T1: Semantic Seg.} & \multicolumn{5}{c}{T2: SN Prediction} & Sparsity & Pre-trained\\
            & mIoU$\uparrow$ & PixelAcc$\uparrow$ & Mean Err$\downarrow$ & Median Err$\downarrow$ & 11.25$\uparrow$ & 22.5$\uparrow$  & 30$\uparrow$& (\%)$\uparrow$\\
            \midrule
            Cross-Stitch~\cite{misra2016cross} & {25.3} & {57.4} & {16.6} & {13.2} & {43.7} & {72.4} & {83.8} & 0 &\xmark\\
            Sluice~\cite{ruder2019latent} & {26.6} & {59.1} & {16.6} & \textbf{13.0} & \textbf{44.1} & {73.0} & {83.9} & 0 &\xmark\\
            DEN~\cite{ahn2019deep} & {26.3} & {58.8} & {17.0} & {14.3} & {39.5} & {72.2} & {84.7} & 0 &\xmark\\
            \method{}(\textit{Static}) & {26.5} & {57.8} & \textbf{16.4} & {13.7} & {41.2} & \textbf{74.1} & \textbf{85.9} & \textbf{90} &\xmark\\
            \method{} (\textit{Dynamic}) & \textbf{28.2} & \textbf{59.2} & 16.5 & {13.5} & \textbf{42.3} & {73.1} & {84.7} & \textbf{90} &\xmark\\
            \bottomrule
        \end{tabular}
    }
    \caption{\method{} semantic segmentation and surface normal prediction results on NYU-v2~\cite{silbermanECCV12} compared to other MTL approaches.}
    \label{table:comp_mtl}
\end{table*}

\begin{table*}[b!]
    \centering
    \resizebox{0.993\textwidth}{!}
    {
        \begin{tabular}{l|cc|ccccc|c|c}
            \toprule
            Model  & \multicolumn{2}{c}{T1: Semantic Seg.} & \multicolumn{5}{c}{T2: SN Prediction} & Sparsity & Pre-trained\\
            & mIoU$\uparrow$ & PixelAcc$\uparrow$ & Mean Err$\downarrow$ & Median Err$\downarrow$ & 11.25$\uparrow$ & 22.5$\uparrow$  & 30$\uparrow$& (\%)$\uparrow$\\
            \midrule
            DeepLab~\cite{chen2017deeplab}{\footnotesize (baseline)} & 27.69 & 58.77 & 16.55 & 14.17 & 39.62 & 73.54 & 86.33 & 0 & N/A \\
            LTH~\cite{frankle2018lottery} & 23.84 & 56.35 & 16.81 & 13.84 & 40.91 & 72.31 & \textbf{84.28} & 30.00 & \cmark\\
            SNIP~\cite{lee2018snip} & 26.57 & 59.85 & 16.91 & 13.55 & 42.01 & 71.72 & 82.01 & 30.00   & \xmark\\
            Random & 25.08 & 55.56 & 17.60 & 14.27 & 40.49 & 70.12 & 81.68 & 30.00 & \xmark\\
            \method{} (\textit{Ours}) & \textbf{28.24} & \textbf{60.33} & \textbf{16.62} & \textbf{13.37} & \textbf{42.98} & \textbf{72.29} & 83.96 & 30.00 &\xmark\\
            \midrule
            RigL~\cite{evci2020rigging} & 24.83 & 57.92 & 16.78 & 14.84 & 37.76 & 72.18 & 86.15 & 30.00 &\xmark\\
            \method{} (\textit{Ours}) & \textbf{28.41} & \textbf{59.77} & \textbf{16.54} & \textbf{13.48} & \textbf{43.42} & \textbf{73.55} & \textbf{86.76} & 30.00 &\xmark\\
            \midrule
            IMP~\cite{han2015deep} & 29.23 & 59.83 & 16.57 & 13.38 & 43.16 & 72.41 & \textbf{84.14} & 30.00 &\cmark\\
            Random & 26.43 & 58.25 & 16.89 & 13.71 & 41.92 & 71.72 & 83.77 & 30.00 &\cmark\\
            \method{} (\textit{Ours}) & \textbf{29.44} & \textbf{59.98} & \textbf{16.56} & \textbf{13.35} & \textbf{43.21} & \textbf{72.25} & 84.06 & 30.00 &\cmark\\
            \bottomrule
        \end{tabular}
    }
    \caption{\method{} semantic segmentation and surface normal prediction results on NYU-v2~\cite{silbermanECCV12} compared to static sparse training, dynamic sparse training, and pre-trained model pruning methods.}
    \label{table:appnyu}
\end{table*}

\setlength{\tabcolsep}{2pt}
\begin{table*}[b!]
    \centering
    \resizebox{0.994\textwidth}{!}
    {
        \begin{tabular}{l|ccc|ccccc|c|c}
            \toprule
            Model  & \multicolumn{3}{c}{T1: Semantic Seg.} & \multicolumn{5}{c}{T2: Depth Prediction} & Sparsity & Pre-trained\\
            & mIoU~$\uparrow$             & PixelAcc~$\uparrow$ & Error~$\downarrow$       & Abs. Error~$\downarrow$ & Rel. Error~$\downarrow$ & $\delta 1.25$ $\uparrow$ & $\delta 1.25^2$ $\uparrow$  & $\delta 1.25^3$ $\uparrow$    & (\%) $\uparrow$      \\
            \midrule
            DeepLab~\cite{chen2017deeplab}{\footnotesize (baseline)} & 42.58 & 74.84 & 0.49 & 0.016 & 0.33 & 74.22 & 88.90 & 94.47 & 0 & N/A\\
            LTH~\cite{frankle2018lottery} & 40.21 & 72.59 & 0.51 & 0.017   & 0.36 & 72.54 & 87.39 & 93.69 & 30.00   & \cmark \\
            SNIP~\cite{lee2018snip} & 41.03 & \textbf{74.65} & 0.51   & 0.018   & 0.36        & 74.80 & \textbf{89.53} & 94.53 & 30.00   & \xmark\\
            Random & 38.17 & 72.77& 0.52   & 0.019   & 0.38 & 72.83 & 83.73 & 92.33 & 30.00   & \xmark \\
            \method{} (\textit{Ours}) & \textbf{42.34} & 74.55 & \textbf{0.49}   & \textbf{0.016}   & \textbf{0.33}        & \textbf{74.91} & 89.22 & \textbf{94.62} & 30.00   & \xmark   \\
            \midrule
            RigL~\cite{evci2020rigging} & 40.68 & 74.40 & 0.51 & 0.018   & 0.36 & 72.47 & \textbf{87.39} & 93.53 & 30.00   & \xmark   \\
            \method{} (\textit{Ours}) & \textbf{42.53} & \textbf{74.82} & \textbf{0.49}   & \textbf{0.016}   & \textbf{0.33}        & \textbf{74.62} & 85.96 & \textbf{93.73} & 30.00   & \xmark   \\
            \midrule
            IMP~\cite{han2015deep} & 42.39  & 72.73 & 0.51   & \textbf{0.016}   & 0.36 & 72.96 & 87.80 & 93.77 & 30.00   & \cmark \\
            Random & 40.14 & 74.41 & 0.52   & 0.018   & 0.39 & 72.38 & 87.90 & 93.85 & 30.00   & \cmark \\
            \method{} (\textit{Ours}) & \textbf{42.47} & \textbf{74.69} & \textbf{0.50} & \textbf{0.016}   & \textbf{0.34}        & \textbf{73.32} & \textbf{88.46} & \textbf{94.37} & 30.00   & \cmark \\
            \bottomrule
        \end{tabular}
    }
    \caption{\method{} semantic segmentation and depth prediction results on Cityscapes~\cite{cordts2016cityscapes} compared to static sparse training, dynamic sparse training, and pre-trained model pruning methods.}
    \label{table:appcity}
\end{table*}

\appendix
\section*{Appendix}
\subsection*{Comparison With MTL Methods}
As mentioned in the Empricial Evaluation Section, \method{} surpasses several dedicated multitask learning approaches despite the high sparsity enforced in our model. In Table \ref{table:comp_mtl}, we show comparison of \method{} in both static and dynamic sparse training setting with several MTL approaches including DEN~\cite{ahn2019deep}, Sluice~\cite{ruder2019latent}, and Cross-Stitch~\cite{misra2016cross} applied on exactly the same model with the same optimization settings. The superiroty of \method{} is clearly observed in the table, demonstrating that \method{} is not only an effective compression approach but also a powerful tool for multitask learning.

\subsection*{Results at Lower Sparsity Levels}
In the Empricial Evaluation Section, we showed the results of \method{} and other pruning and sparse training approaches at high sparsity level($90\%$). Here, in Table \ref{table:appnyu}, \ref{table:appcity}, we show the results at a sparsity of $30\%$, demonstrating the superiority of \method{} at low sparsity level as well. From the table, we can see that \method{} is better across all configurations and evaluation metrics. Moreover, we observed that \method{} achieved lossless compression performance, achieving close or even better performance than the baseline unsparsified model.

\end{document}